\pdfoutput=1

\documentclass[11pt]{article}
\usepackage{graphicx}
\usepackage{float}
\usepackage{hyperref}
\usepackage{url}
\usepackage[]{ACL2023}

\usepackage{times}
\usepackage{latexsym}

\usepackage[T1]{fontenc}

\usepackage[utf8]{inputenc}
\usepackage{microtype}

\usepackage{inconsolata}

\usepackage{enumitem}
\usepackage{graphicx}

\usepackage{booktabs}
\usepackage{amssymb}
\usepackage{amsmath}
\usepackage{adjustbox}
\usepackage{pifont}
\definecolor{lightgreen}{RGB}{0,128,0}
\definecolor{red}{RGB}{255,0,0} 

\title{AI2Apps: A Visual IDE for Building LLM-based AI Agent Applications}

\author{\textbf{Xin Pang}\textsuperscript{1†}, \textbf{Zhucong Li}\textsuperscript{2,4†}, \textbf{Jiaxiang Chen}\textsuperscript{2,4}, \textbf{Yuan Cheng}\textsuperscript{2,3}, \textbf{Yinghui Xu}\textsuperscript{2}, \textbf{Yuan Qi}\textsuperscript{2,3}
\\
$^1$ ContinuAI $^2$ Artificial Intelligence Innovation and Incubation Institute,\\
Fudan University, Shanghai, China\\
$^3$ Shanghai Academy of Artificial Intelligence for Science, Shanghai, China \\
$^4$ School of Computer Science, Fudan University, Shanghai, China \\
pxavdpro@gmail.com, 
\{zcli22, jiaxiangchen23\}@m.fudan.edu.cn\\
\{cheng\_yuan, xuyinghui, qiyuan\}@fudan.edu.cn}
\begin{document}
\maketitle
{
\renewcommand{\thefootnote}{\fnsymbol{footnote}}
\footnotetext[2]{Corresponding author.}
}
\begin{abstract}

We introduce \textbf{AI2Apps}, a Visual Integrated Development Environment (Visual IDE) with full-cycle capabilities that accelerates developers to build deployable LLM-based \textbf{AI} agent \textbf{App}lication\textbf{s}. 
This Visual IDE prioritizes both the \textbf{Integrity} of its development tools and the \textbf{Visuality} of its components, ensuring a smooth and efficient building experience.
On one hand, AI2Apps integrates a comprehensive development toolkit—ranging from a prototyping canvas and AI-assisted code editor to agent debugger, management system, and deployment tools—all within a web-based graphical user interface. On the other hand, AI2Apps visualizes reusable front-end and back-end code as intuitive drag-and-drop components.
Furthermore, a plugin system named AI2Apps Extension (AAE) is designed for \textbf{Extensibility}, showcasing how a new plugin with 20 components enables web agent to mimic human-like browsing behavior. 
Our case study demonstrates substantial efficiency improvements, with AI2Apps reducing token consumption and API calls when debugging a specific sophisticated multimodal agent by approximately 90\% and 80\%, respectively.
The AI2Apps, including an online demo\footnote{\href{https://www.ai2apps.com}{https://www.ai2apps.com}}, open-source code\footnote{\href{https://github.com/Avdpro/ai2apps}{https://github.com/Avdpro/ai2apps}}, and a screencast video\footnote{\href{https://youtu.be/tQTqxk1LzzU}{https://youtu.be/tQTqxk1LzzU}}, is now publicly accessible.
\end{abstract}

\section{Introduction}
The advent of Large Language Models (LLMs) has significantly advanced the technology of AI agents, paving the way for next-generation applications. Nonetheless, developers are in urgent need of a comprehensive full-stack solution to alleviate the repetitive tasks and escalating costs they face.
However, despite achieving remarkable success, existing efforts to develop LLM-based AI agent applications still face their respective limitations.
LLM Operations (LLMOps) platforms often lack integration with engineering-level tools designed for professional developers, thereby limiting the flexibility in programming and debugging \citep{GPTs,LangSmith,AutoGen_Studio,AppBuilder,Coze,AgentScope,Openagents,Dify,Langflow,Flowise,Bisheng}.
Integrated Development Environments (IDEs) fall short in providing sufficient reusable visual components and remain cumbersome and time-consuming throughout the development process. \citep{Promptflow4VScode,Semantic_Kernel4VScode}. 
Software Development Kits (SDKs), serving as the foundation of agent frameworks, are typically integrated into LLMOps or utilized through IDEs \citep{SemanticKernel,AutoGPT,LangChain,autogen,camel,agentverse,app-builder-SDK,BabyAGI,Promptflow,metagpt}.

\begin{figure}
\centering
\includegraphics[width=0.45\textwidth]{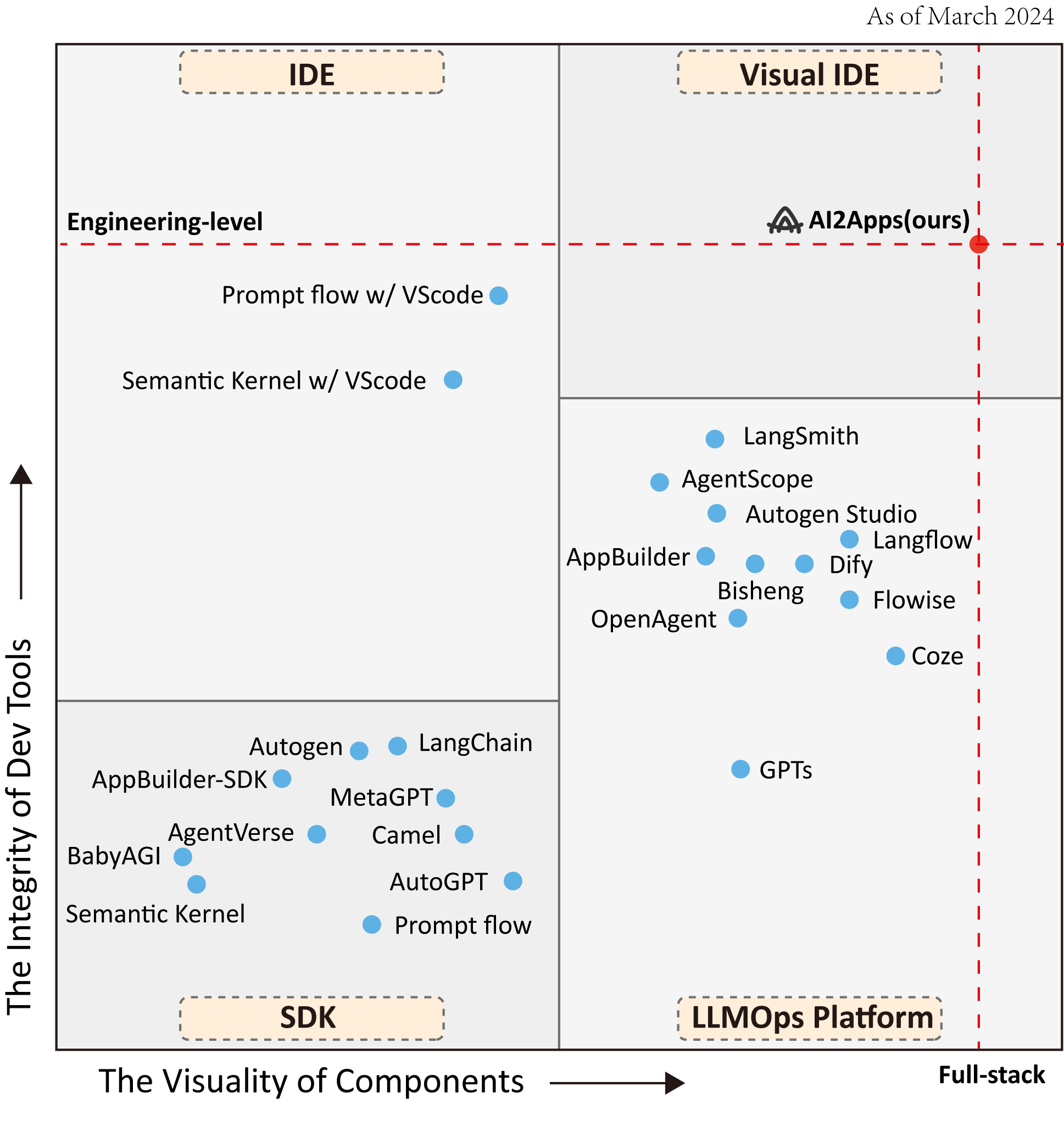}\\
\vspace*{-.2cm}
\caption{The comparison between AI2Apps and existing works on building LLM-based AI agent application is outlined, with the integrity of development tools represented on the vertical axis and the visuality of components indicated on the horizontal axis.}
\vspace*{-.3cm}
\label{compare}
\end{figure}

In response to the aforementioned shortcomings, we introduce \textbf{AI2Apps}, a Visual Integrated Development Environment (Visual IDE) with full-cycle capabilities that empowers developers to efficiently build deployable LLM-based \textbf{AI} agent \textbf{App}lication\textbf{s}. To the best of our knowledge, AI2Apps is the first LLM-based AI agent application development environment that achieves the engineering-level integrity and the full-stack visuality of a Visual IDE, as shown in Figure \ref{compare}.
Its advantages are reflected in:
\begin{enumerate}[leftmargin=*]
  \item \textbf{Integrity}. AI2Apps achieves engineering-level integrity by offering a seamlessly integrated development toolkit within a web-based Graphical User Interface (GUI), featuring a range of tools from a prototyping canvas and AI-assisted code editor to an agent debugger, management system, and deployment tool.
  In design mode, developers can quickly design agents by dragging and dropping components in the \textbf{prototyping canvas}. In code mode, AI2Apps features the \textbf{AI-assisted code editor} designed to help developers write agent application code faster and more consistently. The \textbf{management system} automatically maintains two-way synchronization between the prototyping canvas and the code editor, greatly improving programming efficiency. Then the \textbf{agent debugger} offers topology-based debugging features, including the breakpoint, step run, trace, and GPT mimic. These features allow developers to quickly pinpoint issues and optimize agent performance. Completed AI agents can be packaged as deployable web/mobile applications by \textbf{deployment tool} with one click. They can also be integrated as AI extensions into existing websites/applications with just a few lines of code.
  
  \item \textbf{Visuality}. AI2Apps achieves full-stack visuality by representing multi-dimensional reusable front-end and back-end code as drag-and-drop visual components. 
 AI2Apps provides reusable components oriented toward three dimensions: user interaction, chain, and flow control. The \textbf{user interaction} represents the front-end GUI widgets that facilitate user-agent interaction modes. Relying solely on chat is often not efficient enough to engage with users. Therefore, AI2Apps offers over 50 GUI widgets to support agent development, including menus, buttons, charts, and more. These widgets enable agents to interact with users in a manner akin to real-world applications. The \textbf{chain} represents the backend sequential flow that includes LLMs, prompts, code, agents, and other tools as components. The \textbf{flow control} enables developers to express the logic of applications, incorporating components such as connectors, branches, array loops, summaries, and error handlers. It offers a visual representation of what is traditionally expressed as textual program code.

  \item \textbf{Extensibility}. AI2Apps Extension (AAE) is a plugin extension system specifically designed for AI2Apps. AAE offers developers extensive opportunities to enhance applications by leveraging open technologies as plugins and draggable components. 
  In our screencast video\footnote{\href{https://youtu.be/tQTqxk1LzzU}{https://youtu.be/tQTqxk1LzzU}} we showcase how a new plugin with 20 components enables web agent applications to mimic human-like browsing behavior.

\end{enumerate}

We conduct a case study and find substantial efficiency improvements. With the help of the agent debugger, AI2Apps can reduce token consumption and API calls when debugging a story writing multimodal agent application by approximately 90\% and 80\%, respectively.

\textbf{Our Contributions.} (1) We design a Visual IDE with full-cycle capabilities that accelerates developers to build deployable LLM-based AI agent applications. (2) We implement a plugin extension system for this Visual IDE, offering developers extensive opportunities to enhance applications by freely leveraging open technologies.

\begin{figure*}[!ht]
\centering
\includegraphics[width=1\textwidth]{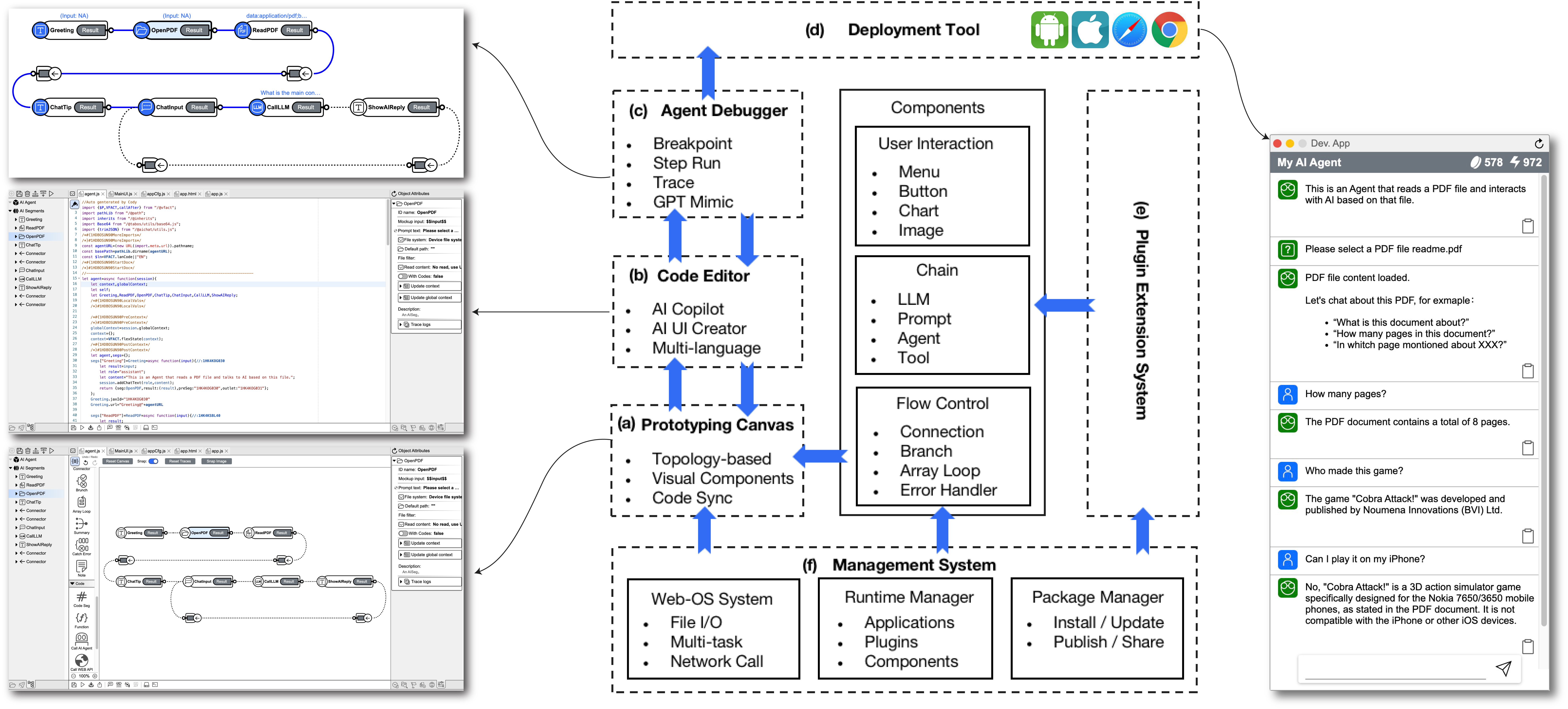}\\
\caption{Architecture of AI2Apps. The left and right sides display screenshots. (a) Prototyping Canvas utilizes built-in components for designing topology diagrams. (b) Code Editor utilizes AI assistance to continue programming the code generated in real-time by the Prototyping Canvas. (c) Agent Debugger pinpoints issues and optimizes agent performance. (d) Deployment Tool releases deployable apps. (e) Plugin Extension System introduces new components. (f) Management System supports the operating environment and resource scheduling.}
\label{architecture}
\end{figure*}

\section{Preliminary}
\subsection{LLM-based AI Agent}
AI agents are artificial entities that are capable of perceiving their surroundings using sensors, making decisions, and then taking actions in response using actuators. LLM-based AI Agent means employing LLMs as the primary component of brain or controller of these agents and expanding their perceptual and action space through strategies such as multimodal perception and tool utilization \citep{xi2023rise,nakano2021webgpt,yao2022react,schick2024toolformer,wei2022chain,kojima2022large,wang2023survey}. They have been applied to various real-world scenarios, such as software development \citep{camel,qian2023communicative,metagpt} and scientific research \citep{boiko2023emergent,bran2023augmenting,boiko2023autonomous}. 

\subsection{Visual IDE}
\textbf{IDEs} are software suites that enhance software development through features like source-code editors, build automation tools, and debuggers. The VScode \citep{VScode} mentioned in Figure \ref{compare} is indeed a widely used IDE software.

\noindent \textbf{Visual IDEs} surpass traditional text-based IDEs by enhancing efficiency, improving code understanding, and facilitating collaboration through user-friendly GUIs, rapid prototyping, and integrated development resources.
The other two concepts mentioned earlier that are easily confused are: \textbf{SDKs} often manifest as Application Programming Interfaces (APIs) or software frameworks and comprise on-device libraries with reusable functions for interfacing with specific programming languages.
\textbf{LLMOps} platforms are specialized tools designed to streamline the deployment, management, and scaling of LLMs. These platforms are crucial for businesses and developers that LLMs for applications such as chatbots, content generation, data analysis, and more. 

\section{The AI2Apps Framework}

AI2Apps is designed for both the integrity of its development tools and the visuality of its components, ensuring a smooth and efficient building experience. It can be organized into six blocks, as shown in Figure \ref{architecture}.

\subsection{Prototyping Canvas}
Prototype design plays a pivotal role in application development by creating interactive models to validate functionality and user experience, ultimately reducing costs and ensuring the delivery of a successful application that meets user needs.
In design mode, developers can quickly design agent logic by dragging and dropping components in the prototyping canvas. 
Its features include:

\noindent \textbf{Topology-based.} The prototyping canvas represents application logic in the form of a topology diagram, thereby breaking down all coupled code into clear and reliable units. Compared to reading complex code line by line, the advantages of AI agent code containing topology diagrams in later maintenance are enormous. It not only provides a clear understanding of the original code's design rationale but also allows for quicker identification of code issues.

\noindent \textbf{Visual Components.} The prototyping canvas visualizes components oriented toward three dimensions: user interaction, chain, and flow control. The user interaction represents the front-end GUI widgets that facilitate user-agent interaction modes. Relying solely on chat is often not efficient enough to engage with users. Therefore, AI2Apps offers over 50 GUI widgets to support agent development, including menus, buttons, charts, and more. These widgets enable agents to interact with users in a manner akin to real-world applications. The chain represents the backend sequential flow that includes LLMs, prompts, code, agents, and other tools as visual components. The flow control enables developers to express the logic of applications, incorporating components such as connectors, branches, array loops, summaries, and error handlers. As a result, developers can easily add and configure them without having to manually write extensive code. This not only enhances development efficiency, but also reduces the likelihood of errors.

\noindent \textbf{Code Sync.} Traditional development approaches struggle to maintain real-time synchronization between the initial design and code implementation, rendering design documents less effective as references during maintenance phases.
AI2Apps' innovative ``Design and Coding at Same Time'' mode guarantees constant alignment between design and implementation, effectively eliminating any discrepancies between the envisioned design and the actual code.

\setlength{\heavyrulewidth}{1.5pt}
\begin{table*}[htbp]
\centering
\begin{adjustbox}{width=1\textwidth}
\begin{tabular}{@{}lcccccccccc@{}}
\toprule
 & \multicolumn{3}{c}{\textbf{Prototyping Canvas}} & \multicolumn{2}{c}{\textbf{Code Editor}} & \multicolumn{3}{c}{\textbf{Agent Debugger}}& \multicolumn{2}{c}{\textbf{Deployment Tool}}\\
\cmidrule(lr){2-4} \cmidrule(l){5-6} \cmidrule(l){7-9} \cmidrule(l){10-11}
\textbf{Name} & \textbf{Topology} & \textbf{Components} & \textbf{Code Sync} &  \textbf{AI Copilot} & \textbf{Multi-lang} & \textbf{Breakpoint} & \textbf{Step Run} & \textbf{Trace} & \textbf{Code} & \textbf{App}\\
\midrule
\textbf{\textit{Visual IDE}:} &  &  &  &  &  &  & & & &\\
\textbf{AI2Apps} (Ours)$^{*}$ & \textcolor{lightgreen}{\checkmark} & \textcolor{lightgreen}{\checkmark} & Two-way & \textcolor{lightgreen}{\checkmark} & \textcolor{lightgreen}{\checkmark} & \textcolor{lightgreen}{\checkmark} & \textcolor{lightgreen}{\checkmark} & \textcolor{lightgreen}{\checkmark} & \textcolor{lightgreen}{\checkmark} & Open \\
\midrule
\textbf{\textit{IDE}:} &  &  &  &  &  &  & & & &\\
Prompt flow w/ VScode \citep{Promptflow4VScode} & \textcolor{lightgreen}{\checkmark} & \textcolor{red}{\ding{55}} & Two-way & \textcolor{red}{\ding{55}} & \textcolor{red}{\ding{55}} & \textcolor{lightgreen}{\checkmark} & \textcolor{lightgreen}{\checkmark} & \textcolor{lightgreen}{\checkmark} & \textcolor{lightgreen}{\checkmark} & Open\\
Semantic Kernel w/ VScode \citep{Semantic_Kernel4VScode} & \textcolor{red}{\ding{55}} & \textcolor{red}{\ding{55}} & \textcolor{red}{\ding{55}} & \textcolor{red}{\ding{55}} & \textcolor{lightgreen}{\checkmark} & \textcolor{lightgreen}{\checkmark} & \textcolor{lightgreen}{\checkmark} & \textcolor{red}{\ding{55}} & \textcolor{lightgreen}{\checkmark} & Open\\
\midrule
\textbf{\textit{LLMOps Platform}:} &  &  &  &  &  &  & & & &\\
GPTs \citep{GPTs} & \textcolor{red}{\ding{55}} & \textcolor{lightgreen}{\checkmark} & \textcolor{red}{\ding{55}} & \textcolor{red}{\ding{55}} & \textcolor{red}{\ding{55}} & \textcolor{red}{\ding{55}} & \textcolor{red}{\ding{55}} & \textcolor{red}{\ding{55}} & \textcolor{red}{\ding{55}} & Closed\\
LangSmith \citep{LangSmith} & \textcolor{red}{\ding{55}} & \textcolor{lightgreen}{\checkmark} & \textcolor{red}{\ding{55}} & \textcolor{red}{\ding{55}} & \textcolor{lightgreen}{\checkmark} & \textcolor{lightgreen}{\checkmark} & \textcolor{lightgreen}{\checkmark} & \textcolor{lightgreen}{\checkmark} & \textcolor{lightgreen}{\checkmark} & Closed\\
Autogen Studio \citep{AutoGen_Studio} & \textcolor{red}{\ding{55}} & \textcolor{lightgreen}{\checkmark} & \textcolor{red}{\ding{55}} & \textcolor{red}{\ding{55}} & \textcolor{red}{\ding{55}} & \textcolor{red}{\ding{55}} & \textcolor{red}{\ding{55}} & \textcolor{red}{\ding{55}} & \textcolor{lightgreen}{\checkmark} & Closed\\
Appbuilder \citep{AppBuilder} & \textcolor{red}{\ding{55}} & \textcolor{lightgreen}{\checkmark} & \textcolor{red}{\ding{55}} & \textcolor{red}{\ding{55}} & \textcolor{lightgreen}{\checkmark} & \textcolor{red}{\ding{55}} & \textcolor{red}{\ding{55}} & \textcolor{red}{\ding{55}} & \textcolor{lightgreen}{\checkmark} & Closed\\
Coze \citep{Coze} & \textcolor{lightgreen}{\checkmark} & \textcolor{lightgreen}{\checkmark} & \textcolor{red}{\ding{55}} & \textcolor{red}{\ding{55}} & \textcolor{red}{\ding{55}} & \textcolor{red}{\ding{55}} & \textcolor{red}{\ding{55}} & \textcolor{red}{\ding{55}} & \textcolor{red}{\ding{55}} & Closed\\
AgentScope$^{*}$ \citep{AgentScope} & \textcolor{red}{\ding{55}} & \textcolor{lightgreen}{\checkmark} & \textcolor{red}{\ding{55}} & \textcolor{red}{\ding{55}} & \textcolor{red}{\ding{55}} & \textcolor{red}{\ding{55}} & \textcolor{red}{\ding{55}} & \textcolor{red}{\ding{55}} & \textcolor{lightgreen}{\checkmark} & Closed\\
OpenAgent$^{*}$ \citep{Openagents} & \textcolor{red}{\ding{55}} & \textcolor{lightgreen}{\checkmark} & \textcolor{red}{\ding{55}} & \textcolor{red}{\ding{55}} & \textcolor{red}{\ding{55}} & \textcolor{red}{\ding{55}} & \textcolor{red}{\ding{55}} & \textcolor{red}{\ding{55}} & \textcolor{red}{\ding{55}} & Closed\\
Dify$^{*}$ \citep{Dify} & \textcolor{red}{\ding{55}} & \textcolor{lightgreen}{\checkmark} & One-way & \textcolor{red}{\ding{55}} & \textcolor{lightgreen}{\checkmark} & \textcolor{red}{\ding{55}} & \textcolor{red}{\ding{55}} & \textcolor{red}{\ding{55}} & \textcolor{lightgreen}{\checkmark} & Closed\\
Langflow$^{*}$ \citep{Langflow} & \textcolor{lightgreen}{\checkmark} & \textcolor{lightgreen}{\checkmark} & One-way & \textcolor{red}{\ding{55}} & \textcolor{red}{\ding{55}} & \textcolor{red}{\ding{55}} & \textcolor{red}{\ding{55}} & \textcolor{red}{\ding{55}} & \textcolor{lightgreen}{\checkmark} & Closed\\
Flowise$^{*}$ \citep{Flowise} & \textcolor{lightgreen}{\checkmark} & \textcolor{lightgreen}{\checkmark} & One-way & \textcolor{red}{\ding{55}} & \textcolor{red}{\ding{55}} & \textcolor{red}{\ding{55}} & \textcolor{red}{\ding{55}} & \textcolor{red}{\ding{55}} & \textcolor{lightgreen}{\checkmark} & Closed\\
Bisheng$^{*}$ \citep{Bisheng} & \textcolor{lightgreen}{\checkmark} & \textcolor{lightgreen}{\checkmark} & One-way & \textcolor{red}{\ding{55}} & \textcolor{red}{\ding{55}} & \textcolor{red}{\ding{55}} & \textcolor{red}{\ding{55}} & \textcolor{red}{\ding{55}} & \textcolor{lightgreen}{\checkmark} & Closed\\
\midrule

\textbf{\textit{SDK}:} &  &  &  &  &  &  & & & &\\
Prompt flow$^{*}$ \citep{Promptflow} & \textcolor{red}{\ding{55}} & \textcolor{red}{\ding{55}} & \textcolor{red}{\ding{55}} & \textcolor{red}{\ding{55}} & \textcolor{red}{\ding{55}} & \textcolor{red}{\ding{55}} & \textcolor{red}{\ding{55}} & \textcolor{red}{\ding{55}} & \textcolor{lightgreen}{\checkmark} & \textcolor{red}{\ding{55}}\\
Semantic Kernel$^{*}$ \citep{SemanticKernel} & \textcolor{red}{\ding{55}} & \textcolor{red}{\ding{55}} & \textcolor{red}{\ding{55}} & \textcolor{red}{\ding{55}} & \textcolor{red}{\ding{55}} & \textcolor{red}{\ding{55}} & \textcolor{red}{\ding{55}} & \textcolor{red}{\ding{55}} & \textcolor{lightgreen}{\checkmark} & \textcolor{red}{\ding{55}}\\
AutoGPT$^{*}$ \citep{AutoGPT} & \textcolor{red}{\ding{55}} & \textcolor{red}{\ding{55}} & \textcolor{red}{\ding{55}} & \textcolor{red}{\ding{55}} & \textcolor{red}{\ding{55}} & \textcolor{red}{\ding{55}} & \textcolor{red}{\ding{55}} & \textcolor{red}{\ding{55}} & \textcolor{lightgreen}{\checkmark} & \textcolor{red}{\ding{55}}\\
BabyAGI$^{*}$ \citep{BabyAGI} & \textcolor{red}{\ding{55}} & \textcolor{red}{\ding{55}} & \textcolor{red}{\ding{55}} & \textcolor{red}{\ding{55}} & \textcolor{red}{\ding{55}} & \textcolor{red}{\ding{55}} & \textcolor{red}{\ding{55}} & \textcolor{red}{\ding{55}} & \textcolor{lightgreen}{\checkmark} & \textcolor{red}{\ding{55}}\\
LangChain$^{*}$ \citep{LangChain} & \textcolor{red}{\ding{55}} & \textcolor{red}{\ding{55}} & \textcolor{red}{\ding{55}} & \textcolor{red}{\ding{55}} & \textcolor{red}{\ding{55}} & \textcolor{red}{\ding{55}} & \textcolor{red}{\ding{55}} & \textcolor{red}{\ding{55}} & \textcolor{lightgreen}{\checkmark} & \textcolor{red}{\ding{55}}\\
Autogen$^{*}$ \citep{autogen} & \textcolor{red}{\ding{55}} & \textcolor{red}{\ding{55}} & \textcolor{red}{\ding{55}} & \textcolor{red}{\ding{55}} & \textcolor{red}{\ding{55}} & \textcolor{red}{\ding{55}} & \textcolor{red}{\ding{55}} & \textcolor{red}{\ding{55}} & \textcolor{lightgreen}{\checkmark} & \textcolor{red}{\ding{55}}\\
MetaGPT$^{*}$ \citep{metagpt} & \textcolor{red}{\ding{55}} & \textcolor{red}{\ding{55}} & \textcolor{red}{\ding{55}} & \textcolor{red}{\ding{55}} & \textcolor{red}{\ding{55}} & \textcolor{red}{\ding{55}} & \textcolor{red}{\ding{55}} & \textcolor{red}{\ding{55}} & \textcolor{lightgreen}{\checkmark} & \textcolor{red}{\ding{55}}\\
AppBuilder-SDK$^{*}$ \citep{app-builder-SDK} & \textcolor{red}{\ding{55}} & \textcolor{lightgreen}{\checkmark} & \textcolor{red}{\ding{55}} & \textcolor{red}{\ding{55}} & \textcolor{lightgreen}{\checkmark} & \textcolor{red}{\ding{55}} & \textcolor{red}{\ding{55}} & \textcolor{red}{\ding{55}} & \textcolor{lightgreen}{\checkmark} & \textcolor{red}{\ding{55}}\\
Camel$^{*}$ \citep{camel} & \textcolor{red}{\ding{55}} & \textcolor{red}{\ding{55}} & \textcolor{red}{\ding{55}} & \textcolor{red}{\ding{55}} & \textcolor{red}{\ding{55}} & \textcolor{red}{\ding{55}} & \textcolor{red}{\ding{55}} & \textcolor{red}{\ding{55}} & \textcolor{lightgreen}{\checkmark} & Closed\\
AgentVerse$^{*}$ \citep{agentverse} & \textcolor{red}{\ding{55}} & \textcolor{red}{\ding{55}} & \textcolor{red}{\ding{55}} & \textcolor{red}{\ding{55}} & \textcolor{red}{\ding{55}} & \textcolor{red}{\ding{55}} & \textcolor{red}{\ding{55}} & \textcolor{red}{\ding{55}} & \textcolor{lightgreen}{\checkmark} & Closed\\
\bottomrule
\end{tabular}%
\end{adjustbox}
\caption{Comparison between AI2Apps and current existing works via the integrity of development tools. The ``$^{*}$'' indicates that the project has been open-sourced. The ``Two-way'' means that the two-way synchronization in real-time between the prototyping canvas and the code editor. The ``One-way'' means only one-way synchronization. The ``Open'' indicates that the generated application belongs to the user's personal copyright and can be freely used. The ``Closed'' indicates that the generated application can only be deployed within the platform itself or encapsulated as an API.
All statistics in the table are collected by March 2024.}
\label{compare_1}
\end{table*}

\setlength{\heavyrulewidth}{1.5pt}
\begin{table}[htbp]
\centering
\scalebox{0.5}{
\begin{tabular}{lccc}
\toprule
\textbf{Name} & \textbf{User Interaction} & \textbf{Chain} & \textbf{Flow Control}\\
\midrule
\textbf{\textit{Visual IDE}:} &  &  &  \\
\textbf{AI2Apps} (Ours)$^{*}$ & \textcolor{lightgreen}{\checkmark} & \textcolor{lightgreen}{\checkmark} & \textcolor{lightgreen}{\checkmark} \\
\midrule
\textbf{\textit{IDE}:} &  &  & \\
PromptFlow w/ VScode \citep{Promptflow4VScode} & \textcolor{red}{\ding{55}} & \textcolor{lightgreen}{\checkmark} & \textcolor{red}{\ding{55}} \\
Semantic Kernel w/ VScode \citep{Semantic_Kernel4VScode} & \textcolor{red}{\ding{55}} & \textcolor{red}{\ding{55}} & \textcolor{red}{\ding{55}} \\
\midrule
\textbf{\textit{LLMOps Platform}:} &  &  & \\
GPTs \citep{GPTs} & \textcolor{red}{\ding{55}} & Limited & \textcolor{red}{\ding{55}} \\
LangSmith \citep{LangSmith} & \textcolor{red}{\ding{55}} & Limited & \textcolor{red}{\ding{55}} \\
Autogen Studio \citep{AutoGen_Studio} & \textcolor{red}{\ding{55}} & Limited & \textcolor{red}{\ding{55}} \\
Appbuilder \citep{AppBuilder} & \textcolor{red}{\ding{55}} & Limited & \textcolor{red}{\ding{55}} \\
Coze \citep{Coze} & \textcolor{red}{\ding{55}} & \textcolor{lightgreen}{\checkmark} & \textcolor{red}{\ding{55}} \\
AgentScope$^{*}$ \citep{AgentScope} & \textcolor{red}{\ding{55}} & Limited & \textcolor{red}{\ding{55}} \\
OpenAgent$^{*}$ \citep{Openagents} & \textcolor{red}{\ding{55}} & Limited & \textcolor{red}{\ding{55}} \\
Dify$^{*}$ \citep{Dify} & \textcolor{red}{\ding{55}} & \textcolor{lightgreen}{\checkmark} & \textcolor{red}{\ding{55}} \\
Langflow$^{*}$ \citep{Langflow} & \textcolor{red}{\ding{55}} & \textcolor{lightgreen}{\checkmark} & \textcolor{red}{\ding{55}} \\
Flowise$^{*}$ \citep{Flowise} & \textcolor{red}{\ding{55}} & \textcolor{lightgreen}{\checkmark} & \textcolor{red}{\ding{55}} \\
Bisheng$^{*}$ \citep{Bisheng} & \textcolor{red}{\ding{55}} & \textcolor{lightgreen}{\checkmark} & \textcolor{red}{\ding{55}} \\
\midrule

\textbf{\textit{SDK}:} &  &  &  \\
Prompt flow$^{*}$ \citep{Promptflow}  & \textcolor{red}{\ding{55}} & \textcolor{red}{\ding{55}} & \textcolor{red}{\ding{55}} \\
Semantic Kernel$^{*}$ \citep{SemanticKernel} & \textcolor{red}{\ding{55}} & \textcolor{red}{\ding{55}} & \textcolor{red}{\ding{55}} \\
AutoGPT$^{*}$ \citep{AutoGPT} & \textcolor{red}{\ding{55}} & \textcolor{red}{\ding{55}} & \textcolor{red}{\ding{55}} \\
BabyAGI$^{*}$ \citep{BabyAGI} & \textcolor{red}{\ding{55}} & \textcolor{red}{\ding{55}} & \textcolor{red}{\ding{55}} \\
LangChain$^{*}$ \citep{LangChain} & \textcolor{red}{\ding{55}} & \textcolor{red}{\ding{55}} & \textcolor{red}{\ding{55}} \\
Autogen$^{*}$ \citep{autogen} & \textcolor{red}{\ding{55}} & \textcolor{red}{\ding{55}} & \textcolor{red}{\ding{55}} \\
MetaGPT$^{*}$ \citep{metagpt} & \textcolor{red}{\ding{55}} & \textcolor{red}{\ding{55}} & \textcolor{red}{\ding{55}} \\
AppBuilder-SDK$^{*}$ \citep{app-builder-SDK} & \textcolor{red}{\ding{55}} & \textcolor{red}{\ding{55}} & \textcolor{red}{\ding{55}} \\
Camel$^{*}$ \citep{camel} & \textcolor{red}{\ding{55}} & \textcolor{red}{\ding{55}} & \textcolor{red}{\ding{55}} \\
AgentVerse$^{*}$ \citep{agentverse} & \textcolor{red}{\ding{55}} & \textcolor{red}{\ding{55}} & \textcolor{red}{\ding{55}} \\
\bottomrule
\end{tabular}
}
\caption{Comparison between AI2Apps and current existing works via the  visuality of components. The ``$^{*}$'' indicates that the project has been open-sourced. The ``Limited'' means that the chain cannot be visualized in an intuitive topology diagram. All statistics in the table are collected by March 2024.}
\label{compare_2}
\end{table}

\subsection{Code Editor}
A code editor is a software tool used by developers to write and edit code for software development projects. In code mode, AI2Apps features the AI-assisted code editor designed to help developers write high-quality application code. Its features include:

\noindent \textbf{AI Copilot.} Our code editor comes with an AI copilot that assists users in generating subsequent code at the cursor position based on context or rewriting the entire document directly. We allow users to freely modify the topology of the AI copilot to enhance its functionality, or users can integrate an external AI copilot through APIs.

\noindent \textbf{AI UI Creator.} The AI UI creator can generate standardized UI code that users need through a conversational interface, utilizing over 50 GUI widgets, and display it in the prototyping canvas.

\noindent \textbf{Multi-language.} The code editor supports programming languages in any form, including JavaScript, Python, and others, enabling developers from any programming background to freely integrate their own code.

\subsection{Agent Debugger}
The design concept of the agent debugger draws inspiration from the debugging functionality found in general-purpose IDEs. Unlike traditional IDEs, which focus on debugging code line by line, the agent debugger is specifically tailored to follow the trajectory of topology diagrams.
Its features include:

\noindent \textbf{Breakpoint.} Setting breakpoints allows pausing execution at specific locations within the topology diagram. This enables developers to inspect the program's state at particular time points, including variable values, memory status, and the program's execution path. Setting breakpoints is an effective troubleshooting method, especially when dealing with complex errors and performance issues.

\noindent \textbf{Step Run.} The step run feature allows developers to execute the topology diagram node by node or to jump into a subgraph within a specific agent, enabling a more detailed examination of the execution flow and the state at various time points. This method is particularly useful for understanding the execution path of the topology diagram, identifying logic errors, and inspecting variable changes.

\noindent \textbf{Trace.} The trace feature visually represents the flow trajectory of data variables on the topology diagram. Users can save and download trace logs in JSON format for subsequent analysis of agents.

\noindent \textbf{GPT Mimic.} During the phase of invoking external LLM APIs, users can set up GPT mimic. When the predefined conditions are met, the API call will no longer go through the network but will directly return the results set by the GPT mimic. This approach can significantly reduce token costs and improve debugging efficiency.

\subsection{Deployment Tool}
Most existing mainstream development platforms develop applications that heavily rely on the platform's own runtime environment, hindering the development of AI agent applications. However, AI2Apps, as a Visual IDE, integrates multiple deployment tools and supports users in directly building externally deployable applications. Completed AI agents can be packaged as standalone web/mobile apps. They can also be integrated as AI extensions into existing websites/apps with just a few lines of code.

\subsection{Plugin Extension System}
AI2Apps Extension (AAE) is a plugin extension system specifically designed for AI2Apps. AAE offers developers extensive opportunities to enhance applications by leveraging open technologies as plugins and draggable components.
AAE features reusable components that allow developers to combine existing ones or create new ones. Developers can share their custom components by publishing them as packages, thereby broadening AI2Apps' functionality through the AAE system.

In our screencast video\footnote{\href{https://youtu.be/tQTqxk1LzzU}{https://youtu.be/tQTqxk1LzzU}} we showcase how a new plugin with 20 components enables web agent applications to mimic human-like browsing behavior. Through web extension components, AI2Apps is equipped with comprehensive control over web pages, enabling actions such as opening/switching pages, reading page content, filling/modifying page content, simulating user behavior, and more.

\subsection{Management System}
Management System provides various underlying functional support for AI2Apps, including:

\noindent \textbf{Web-OS System.} Based on web API technology, it provides a complete set of desktop operating system functions for AI2Apps. Examples include file system, multi-task support, network call. It offers users a development experience that is comparable to native operating systems, yet more convenient and secure.

\noindent \textbf{Runtime Manger.} It provides specialized runtime foundational functions for AI-oriented applications, including: application management/scheduling, plugin extension/control, integration of AI functionality components, etc.

\noindent \textbf{Package Manager.} It extends the capabilities of AI2Apps through packages, which include services for building, publishing, sharing packages, as well as installation and upgrades.

\section{Case Study and Usage}
\begin{figure}
\centering
\includegraphics[width=0.4\textwidth]{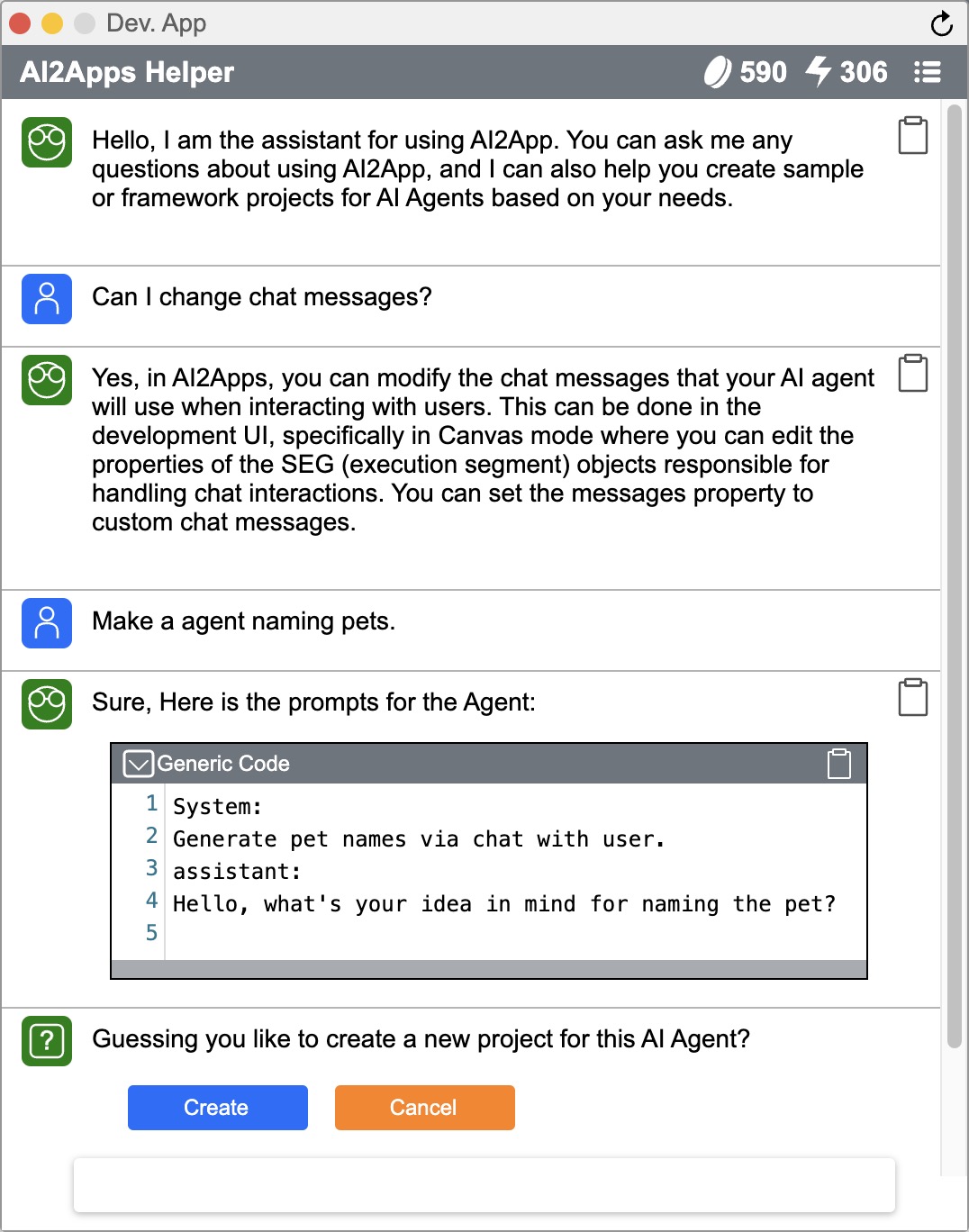}\\
\vspace*{-.2cm}
\caption{Screenshot of our usage assistant built by AI2Apps.}
\vspace*{-.3cm}
\label{debug}
\end{figure}

\subsection{Case Study}
We conduct a case study and find substantial efficiency improvements.
In this case, we initially utilized AI2Apps to build a complex Agent application for writing short stories. It allows specifying drawing styles, precise unlimited-length writing with the ability to include covers and illustrations, and possesses powerful editing capabilities for articles. It supports paragraph-by-paragraph modifications, including content and images, and provides the option to choose whether AI assistance is used during editing. We conducted four sets of controlled experiments by clearing all prompts from the agent and having eight volunteers attempt to restore the functionality of the original agent, with half using the agent debugger and the other half without it.
With the help of the agent debugger, AI2Apps reduced token consumption and API calls when debugging a story writing multimodal agent application by approximately 90\% and 80\%, respectively. The agent debugger user interface is shown in Figure \ref{debug}. This agent project can be accessed at \href{https://github.com/Avdpro/StoryWriter}{https://github.com/Avdpro/StoryWriter}

\subsection{Usage}
Given the extensive functionality of AI2Apps, this paper is unable to offer detailed usage guidelines within its confined scope. For more information about our usage example, please refer to \href{https://github.com/Avdpro/ai2apps}{https://github.com/Avdpro/ai2apps}. We've built a usage assistant using AI2Apps, as shown in Figure \ref{debug}, allowing users to get started with AI2Apps by simply running it at \href{https://www.ai2apps.com}{https://www.ai2apps.com}.

\section{Comparison with Related Works}
In Figure \ref{compare}, Table \ref{compare_1} and Table \ref{compare_2}, we present a detailed comparison of existing works through the perspectives of the integrity of development tools and the visuality of components. To the best of our knowledge, AI2Apps is the first LLM-based AI agent application development environment that achieves the engineering-level integrity and the full-stack visuality of a Visual IDE.
Visual IDEs can integrate SDKs like traditional IDEs, and they can also be integrated into LLMOps platforms. Therefore, the open-source release of AI2Apps effectively fills the current gap in development technology, facilitating the advancement of AI agent applications.

\section{Conclusion}
We introduced AI2Apps, the first Visual IDE with full-cycle capabilities that empowers developers to efficiently build deployable LLM-based AI agent Applications. 
AI2Apps offers the engineering-level development tools and the full-stack visual components.
It features a web-based interface with tools such as a prototyping canvas, an AI-powered code editor, a agent debugger, a management system, and deployment tools, alongside intuitive drag-and-drop components for code visualization. Additionally, we implemented a plugin extension system for this Visual IDE, offering developers extensive opportunities to enhance applications by freely leveraging open technologies. We conducted a case study and find substantial efficiency improvements. With the help of the agent debugger, AI2Apps can reduce token consumption and API calls when debugging a story writing multimodal agent application by approximately 90\% and 80\%, respectively.

\section*{Limitations}
As a Visual IDE, AI2Apps still cannot  fully match the flexibility of traditional IDEs in app development, nor does it offer the operational capabilities of LLMOps platforms for deployed applications. However, its development convenience is noteworthy. we warmly welcome LLMOps platforms to integrate this work to enhance their offerings, thus jointly promoting related academic research.

\section*{Ethics Statement}
(1) This material is the authors’ own original work,
which at this stage of project development has not
been previously published elsewhere. 
(2) The paper is not currently being considered for publication
elsewhere. 
(3) The paper reflects the authors’ own research and analysis in a truthful and complete manner. 
(4) Our work does not contain identity characteristics. It does not harm anyone. The eight participants in the case study part are volunteers recruited from students majoring in engineering. Before the case study experiments, all participants are provided with detailed guidance in both written and oral form. The only recorded user-related information is usernames, which are anonymized and used as identifiers to mark different participants.
(5) AI2Apps is designed to help users to build deployable AI agent applications. 
(6) Our work does not involve LLM training or fine-tuning; we only use publicly available APIs permitted for research purposes. Therefore, there are no data-related risks associated with our approach.

\bibliography{custom}
\bibliographystyle{acl_natbib}

\end{document}